\documentclass{article}
\usepackage{ijcai22}

\usepackage{times}
\usepackage{soul}
\usepackage{url}
\usepackage[hidelinks]{hyperref}
\usepackage[utf8]{inputenc}
\usepackage[small]{caption}
\usepackage{graphicx}
\usepackage{amsmath}
\usepackage{amsthm}
\usepackage{booktabs}
\usepackage{algorithm}
\usepackage{algorithmic}
\usepackage{setspace}
\usepackage{amssymb,amsfonts}
\usepackage[dvipsnames]{xcolor}
\usepackage[makeroom]{cancel}
\definecolor{orange}{HTML}{B24A17}
\definecolor{myblue}{HTML}{177FB2}

\usepackage{balance}

\title{
Summarising and Comparing Agent Dynamics\\with Contrastive Spatiotemporal Abstraction
}

\author{
    Tom Bewley\and
    Jonathan Lawry\and
    Arthur Richards
    \affiliations
    University of Bristol, United Kingdom
    \emails
    \{tom.bewley,\ l.lawry,\ arthur.richards\}@bristol.ac.uk
}

\begin{document}

\vspace{-1cm}
\maketitle
\vspace{-1cm}

\begin{abstract}
We introduce a data-driven, model-agnostic technique for generating a human-interpretable summary of the salient points of contrast within an evolving dynamical system, such as the learning process of a control agent. It involves the aggregation of transition data along both spatial and temporal dimensions according to an information-theoretic divergence measure. A practical algorithm is outlined for continuous state spaces, and deployed to summarise the learning histories of deep reinforcement learning agents with the aid of graphical and textual communication methods. We expect our method to be complementary to existing techniques in the realm of agent interpretability. 
\end{abstract}

\section{INTRODUCTION} \label{sec:intro}

\begin{figure*}[t]
\centering
\includegraphics[width=\textwidth]{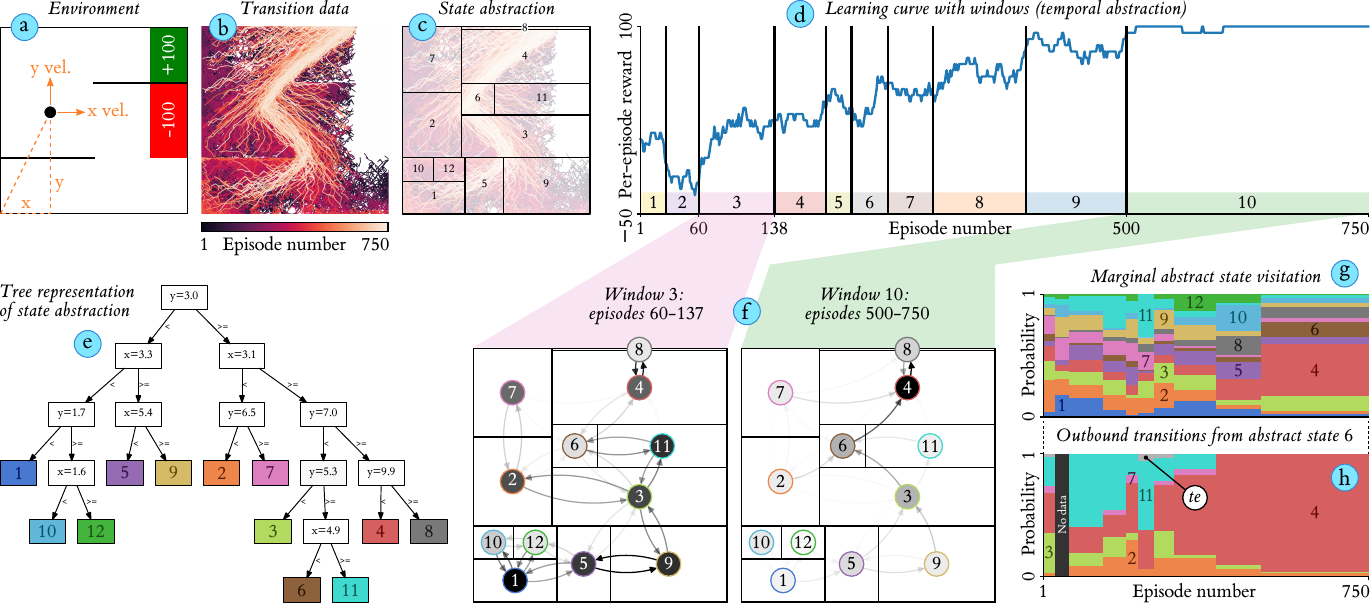}
\caption{
\small 
Using our method
to summarise an RL agent's learning in a maze (a) with positional state dimensions $x,y$. Given transition data from $750$ learning episodes (b), we partition first along the state dimensions to create $12$ abstract states (c), then temporally to create $10$ time windows (d), in both cases greedily maximising the Jensen-Shannon divergence among the set of abstract transition distributions. We can explore the representation by displaying the state abstraction as a tree diagram (e) and the abstract Markov model of each window's dynamics as a transition graph (f). We can also represent the time evolution explicitly using visualisations of the visitation probabilities for all abstract states (g) and conditional probabilities of outbound transitions from a single abstract state (h). \textit{Note: $te=$ episode termination.}\vspace{-0.15cm}
}
\label{fig:results_holonav}
\end{figure*}

In response to the growing complexity of machine learning, the field of explainable artificial intelligence (XAI) aims to ``open the black box" of data-driven learning algorithms, providing human-interpretable summaries of the patterns, biases and defects in their behaviour \cite{adadi2018peeking}. One class of learning algorithm that has received comparatively little attention from the XAI community are those that
learn control policies for agents in dynamic environments, such as reinforcement learning (RL) \cite{sutton2018reinforcement}. 
This may be because control problems have dependencies that make localised modelling or explanation difficult. The non-i.i.d. nature of the data, the feedback between an agent's past experience and current policy, the use of random exploration, and the lack of a clear notion of predictive accuracy all complicate causal disentanglement and make it hard to generate coherent, self-contained explanations. Many existing methods circumvent these complexities by interpreting an agent's action or belief state at a single point in time and state space, borrowing techniques from the wider XAI literature such as saliency maps \cite{huber2019enhancing},
counterfactuals \cite{bewley2021tripletree},
and rule-based distillation \cite{bastani2018verifiable}.
Such single-timestep explanations produce valuable insight, but overlook the dynamics that differentiate control from other learning domains. Other approaches include temporal information by synthesising informative trajectory demonstrations \cite{amir2018highlights,huang2019enabling}, although generalisable insight relies on a user's correct interpretation of the particular demonstrations seen. A complementary direction would be to extract global summaries of agent dynamics over two timescales: (1) short-term state transitions (\textit{``when you go here, what happens next?"}) and (2) long-term policy evolution (\textit{``what did you do in the past, and when, how and why did that change?"}). 

We introduce a framework for interpretable, algorithm-agnostic modelling of an agent's dynamics at both timescales. We harness transition data generated through the learning process to partition the state space into regions called abstract states, then approximate the nonstationary behaviour as a sequence of Markov chains over abstract states, each associated with a time window. Both abstract states and window boundaries are defined to maximise an information-theoretic divergence measure, so that the set of abstract transition probabilities captures the salient changes in behaviour in a heavily-compressed form. This in turn allows us to take a contrastive perspective on explanation, favoured in both the psychological \cite{lipton1990contrastive}
and computational \cite{miller2018contrastive} literature, by summarising the most significant differences in the probabilities of two or more abstract Markov chains. While abstract transition models have been used before as a medium for agent behaviour explanation \cite{zahavy2016graying,topin2019generation,davoodi2021feature,mccalmon2022caps}, our work is the first to handle multiple or nonstationary policies, and develop principled theories and algorithms based on the notion of contrastive explanation.

Figure \ref{fig:results_holonav} shows the application of our method to the learning history of a deep RL agent in a maze. After outlining our problem statement and methods in Sections 2 and 3, we return in Section 4 to discuss these results in detail, alongside those from a more complex and high-dimensional control task.

\section{PROBLEM STATEMENT} \label{sec:problem}

Consider a sequence of $k$ Markov chains over a common state space $\mathcal{S}$, i.e. $\mathcal{P}=(P_1,...,P_k)$, where $P_i:\mathcal{S}\rightarrow\Delta(\mathcal{S}),\ \forall i\in\{1,...,k\}$ and $\Delta(\cdot)$ is the set of all distributions over a set.

\paragraph{Remark} \textit{Initially, we can be agnostic to the generative origins of $\mathcal{P}$ and the relationships between the chains (e.g. ordering may or may not be important). Moreover, our abstraction approach can in principle be applied to any kind of state space: discrete, continuous or mixed. In the algorithms and experiments below, we assume a continuous space $\mathcal{S}=\mathbb{R}^D$.
}

Now suppose we do not have direct access to $\mathcal{P}$, but can collect a transition dataset $\mathcal{D}$ comprising elements of the form $(i,s,s')$, where $i$ is the index of one of the Markov chains and $s,s'$ is a pair of successive states sampled from its induced distribution.\footnote{For ergodic Markov chains, this is the stationary distribution. For absorbing chains \cite{kemeny1976finite}, let this be the distribution over \textit{pre-absorption} transitions, where $s$ is a transient state. 
} Our goal is to use $\mathcal{D}$ to facilitate intuitive
comparison
of the underlying dynamics in $\mathcal{P}$. To do so, we propose to give the data a more compact, interpretable representation through the use of \textit{state abstraction}. A state abstraction is a set $\mathcal{X}=\{x_1,...,x_m\}$, whose $m$ elements (called abstract states) partition $\mathcal{S}$, i.e. $\bigcup_{x\in\mathcal{X}}x=\mathcal{S}$ and $x\neq x'\implies x\cap x'=\emptyset,\ \forall x,x'\in\mathcal{X}$. 
Every $s\in\mathcal{S}$
is therefore a member of exactly one $x\in\mathcal{X}$, which we denote by $s\in x$. 
For any given $\mathcal{X}$, the transition data $\mathcal{D}$ have an alternative representation as a $k\times m\times m$ array of abstract transition counts $N^\mathcal{X}$, where $\forall i\in\{1,...,k\},x\in\mathcal{X},x'\in\mathcal{X}$, the count is computed by
\begin{equation} \label{eq:counts}
N^\mathcal{X}_{i,x,x'}=|\{(j,s,s')\in\mathcal{D}:j=i\land s\in x\land s'\in x'\}|.
\end{equation}

Let $N^\mathcal{X}_i$ be the $k\times k$ ``slice" of $N^\mathcal{X}$ for chain $i$. Normalising $N^\mathcal{X}_i$ by its grand sum yields an empirical joint distribution over abstract transitions $J^\mathcal{X}_i$, and separately normalising by row gives a conditional distribution $P^\mathcal{X}_i$. The latter is of particular interest as it defines a Markov chain over the discrete space $\mathcal{X}$, which may be compact enough to visualise and comprehend in its entirety, and to which we may apply established methods to analyse properties such as convergence, stability and cycles. Our hypothesis is that provided we always remain aware of the epistemic relationship between $P^\mathcal{X}_i$ and $P_i$ (the former is a compressed model of the latter, estimated from limited data) and if $\mathcal{X}$ is carefully constructed, analysis of the abstract Markov chains can provide meaningful and accurate insight into the underlying dynamics. 

What do we mean by the term ``carefully constructed" in the preceding paragraph? Clearly, there is a tradeoff around the abstraction size $m$, with the extrema $\mathcal{X}=\{\mathcal{S}\}$ ($\therefore m=1$) and $\mathcal{X}=\mathcal{S}$ ($\therefore m=|\mathcal{S}|$) respectively providing a degenerate model or zero interpretability gain. 
But beyond this, we can reasonably expect some $m$-sized state abstractions to be more effective than others for the purpose of capturing salient features of the underlying dynamics. 
We investigate this hypothesis by postulating an objective for constructing $\mathcal{X}$.

\subsection{The Contrastive Objective} 

According to several influential theories  \cite{hesslow1988problem,van1988pragmatic,lipton1990contrastive}, explanations 
are constructed by searching for contrasts with alternative cases. This in turn implies that a common representation of multiple phenomena has explanatory value to the extent that it preserves information about their mutual differences, while ignoring irrelevant details. Motivated by this reasoning, we propose that a useful abstraction $\mathcal{X}$ is compact (small $m$) while nonetheless maximising our ability to \textit{discriminate between chains $i\in\{1,...,k\}$ based on their abstract transition probabilities}. This is practically facilitated by placing the boundaries of abstract states at points of divergence between the chains, thus disentangling their transitions.

Let us consider the sequence of abstract models, defined by the joint probabilities $J^\mathcal{X}_1,...,J^\mathcal{X}_k$, as generative distributions from which abstract transitions can be sampled. Applying Bayes' rule, the posterior probability of a chain $i$ given a transition $x,x'$ sampled from its abstract model is
\begin{equation} \label{eq:posterior}
\text{Pr}(i|x,x')=\frac{\rho_iJ^\mathcal{X}_{i,x,x'}}{\sum_{j=1}^k\rho_jJ^\mathcal{X}_{j,x,x'}},
\end{equation}
where the vector $\rho\in[0,1]^k$, $\sum_{i=1}^k\rho_i=1$ specifies a prior weighting over chains. Taking the prior-weighted expectation of the log posterior over all chains and transitions, we obtain:
\begin{multline}\label{eq:jsd_deriv}
\underset{i,x,x'}{\mathbb{E}}\left[\log \text{Pr}(i|x,x')\right]=\\
\sum_{i=1}^k\rho_i\sum_{x\in\mathcal{X}}\sum_{x'\in\mathcal{X}}J^\mathcal{X}_{i,x,x'}\left[\log \frac{\rho_iJ^\mathcal{X}_{i,x,x'}}{\sum_{j=1}^k\rho_jJ^\mathcal{X}_{j,x,x'}}\right]\\
=\sum_{i=1}^k\rho_i\log \rho_i\cancelto{=1}{\sum_{x\in\mathcal{X}}\sum_{x'\in\mathcal{X}}J^\mathcal{X}_{i,x,x'}}\\
+\sum_{i=1}^k\rho_i\sum_{x\in\mathcal{X}}\sum_{x'\in\mathcal{X}}J^\mathcal{X}_{i,x,x'}\log J^\mathcal{X}_{i,x,x'}\\
-\sum_{x\in\mathcal{X}}\sum_{x'\in\mathcal{X}}\Big(\sum_{i=1}^k\rho_iJ^\mathcal{X}_{i,x,x'}\Big)\log\Big( \sum_{j=1}^k\rho_jJ^\mathcal{X}_{j,x,x'}\Big)\\
=-\mathcal{H}(\rho)-\sum_{i=1}^k\rho_i\mathcal{H}(J^\mathcal{X}_i)+\mathcal{H}\Big(\sum_{i=1}^k\rho_iJ^\mathcal{X}_i\Big)\\
=\text{JSD}(J^\mathcal{X}|\rho)-\mathcal{H}(\rho).
\end{multline}

$\mathcal{H}$ is entropy, and $\text{JSD}$ denotes the multi-distribution generalisation of the Jensen-Shannon divergence \cite{lin1991divergence}, which is equivalent to the mutual information between the prior-weighted mixture distribution $\sum_{i=1}^k\rho_iJ^\mathcal{X}_i$ and the indicator variable $i$. Intuitively, Equation \ref{eq:jsd_deriv} tells us that chain discrimination is facilitated by an abstraction that yields abstract transition probabilities $J^\mathcal{X}_1,...,J^\mathcal{X}_k$ which are as mutually-divergent as possible, and also by minimising $\mathcal{H}(\rho)$, although this second term is not relevant here since we take $\rho$ to be a fixed parameter specifying our level of interest in each chain. 

We therefore propose the following \textit{contrastive abstraction} objective. Given a set of state abstractions $\mathbb{X}$ constructible by some well-defined algorithm, the objective is
$$
\underset{\mathcal{X}\in\mathbb{X}}{\text{argmax}}\Big[\ \underset{i,x,x'}{\mathbb{E}}\log \text{Pr}(i|x,x')-\alpha (m-1)\ \Big]
$$
\vspace{-0.2cm}
\begin{equation} \label{eq:objective}
=\underset{\mathcal{X}\in\mathbb{X}}{\text{argmax}}\Big[\ \text{JSD}(J^\mathcal{X}|\rho)-\alpha (m-1)\ \Big].
\end{equation}
Here, $\mathcal{H}(\rho)$ is ignored because it is independent of $\mathcal{X}$, and the second term (parameterised by $\alpha>0$) is a complexity regulariser to incentivise smaller abstractions. This term is required to ensure the objective is not maximised by the trivial solution $\mathcal{X}=\mathcal{S}$. In expectation, the $\text{JSD}$ term increases monotonically but sublinearly with $m$ (see Appendix A for a simple theoretical analysis and empirical demonstration), and $\alpha (m-1)$ is evidently linear, so we should expect their difference to be maximised at an intermediate abstraction size determined by $\alpha$ (larger $\alpha$ favours smaller $m$).

\subsection{Temporal Abstraction}

The preceding discussion concerns a $k$-chain sequence $\mathcal{P}=(P_1,...,P_k)$.
A special case of this setup is when the sequence is the product of a contiguous, temporally-evolving process: a \textit{nonstationary} Markov chain. Suppose that $k$ is very large, and that the dataset $\mathcal{D}$ contains a small number of samples from each $P_i$ (perhaps as few as $1$). Regardless of our choice of state abstraction $\mathcal{X}$, the result will be an array of joint probabilities $J^\mathcal{X}$ that is not only large (hampering interpretability) but also sparse, meaning the abstract Markov chains will have such high variance as to be practically useless as models of $\mathcal{P}$. 

In many real-world dynamical systems, however, changes are incremental, so it is reasonable to assume approximate stationarity over short timescales. We therefore propose to reduce both the size and the sparsity of $J^\mathcal{X}$ by applying a second layer of \textit{temporal abstraction}, which approximates the dynamics as piecewise constant within $n\ll k$ disjoint and exhaustive temporal windows $\mathcal{T}=\{\tau_1,...,\tau_n\}$, where each $\tau_w$ is parameterised by lower and upper bounds $l_w,u_w$, and the following constraints hold: $l_1=1$; $u_n=k+1$; $l_w<u_w,\ \forall w\in\{1,...,n\}$; $u_w=l_{w+1},\ \forall w\in\{1,...,n-1\}$.
These can be used to aggregate $J^\mathcal{X}$ along its first axis to form a smaller and less sparse $n\times m\times m$ array $J^{\mathcal{X},\mathcal{T}}$, where $\forall w\in\{1,...,n\},x\in\mathcal{X},x'\in\mathcal{X}$,
\begin{equation}
\vspace{-0.15cm}
J_{w,x,x'}^{\mathcal{X},\mathcal{T}}=\frac{1}{\rho^\mathcal{T}_w}\sum_{i=l_w}^{u_w-1}\rho_iJ^\mathcal{X}_{i,x,x'},
\end{equation}
and $\rho^\mathcal{T}_w=\sum_{i=l_w}^{u_w-1}\rho_i$ aggregates the elements of the prior vector for each window.
By varying both $n$ and the 
window boundaries in $\mathcal{T}$,
we obtain different sequences of spatiotemporally-abstracted transition probabilities, even if $\mathcal{X}$ remains constant. Intuitively, a temporal abstraction is informative for contrastive analysis if it 
contains windows
of behaviour which are internally consistent, but between which there are significant differences.
We suggest
that this can be achieved using an \textit{identical objective} to that of the previous section: maximising the Jensen-Shannon divergence between the joint transition distributions of the 
windows.
We again wish to incentivise compact temporal abstractions (small $n$), so add another regularisation term to the objective:
\begin{equation} \label{eq:objective_with_windows}
\underset{\mathcal{X}\in\mathbb{X},\mathcal{T}\in\mathbb{T}}{\text{argmax}}\Big[\ \text{JSD}(J^{\mathcal{X},\mathcal{T}}|\rho^\mathcal{T})-\alpha (m-1)-\beta(n-1)\ \Big],
\end{equation}
where $\mathbb{T}$ is the set of temporal abstractions constructible by some algorithm, and $\beta>0$ is a hyperparameter. Theoretical and empirical results again indicate that $\text{JSD}$ is sublinear in $n$ (Appendix A). As with $m$, this means that the objective will be maximised at a finite value of $n$ determined by $\beta$.

\section{ABSTRACTION ALGORITHMS} \label{sec:algos}

We now present algorithms for contrastive spatiotemporal abstraction given transition data $\mathcal{D}$ induced by a Markov chain sequence $\mathcal{P}$ in a continuous state space $\mathcal{S}=\mathbb{R}^D$. The central idea is to recursively partition $\mathcal{S}$ with axis-aligned hyperplanes, yielding hyperrectangular abstract states whose simple geometry aids interpretability. The top-down partitioning approach imposes a structural bias towards compact abstractions, and means the algorithms need only consider solutions up to some acceptable maximum size. The core algorithm, designed to optimise the objective in Equation \ref{eq:objective} and used in the case of $k$ distinct Markov chains, is referred to as \texttt{CSA}. The extension for temporal abstraction of a nonstationary Markov chain is \texttt{CSTA}, which first calls \texttt{CSA} before running an algorithmically-similar temporal partitioning loop to find windows that maximise the extended objective in Equation \ref{eq:objective_with_windows}. In this section, we provide an overview of the approach with examples and compact pseudocode. Annotatated 
pseudocode
and further algorithmic details are given in Appendix B.

\subsection{\texttt{CSA} (\textit{Constrastive State Abstraction})} 

Suppose we have an existing abstraction $\mathcal{X}$, consisting of $m$ hyperrectangles in the state space $\mathcal{S}=\mathbb{R}^D$, and have precomputed the joint abstract transition probabilities $J^\mathcal{X}$ by equation \ref{eq:counts} and normalisation.
Now consider splitting an abstract state $x\in\mathcal{X}$ into two by placing a hyperplane at a location $c\in\mathbb{R}$ along state dimension $d\in\{1,...,D\}$. All transitions into, out of and within the abstract state will be redistributed between the two ``child" abstract states. This can be represented by an enlarged transition array with shape $k\times(m+1)\times(m+1)$, denoted by $J^{\mathcal{X}\rightarrow[xdc]}$. The splitting operation increases the quality of the abstraction according to the objective in equation \ref{eq:objective} by an amount equal to the increase in JSD (always $+$ve), minus the regularisation parameter $\alpha$:
\begin{equation} \label{eq:greedy}
\delta(\mathcal{X}\rightarrow[xdc])=\text{JSD}(J^{\mathcal{X}\rightarrow[xdc]}|\rho)-\text{JSD}(J^\mathcal{X}|\rho)-\alpha.
\end{equation}

\begin{figure}[h!]
\centering
\includegraphics{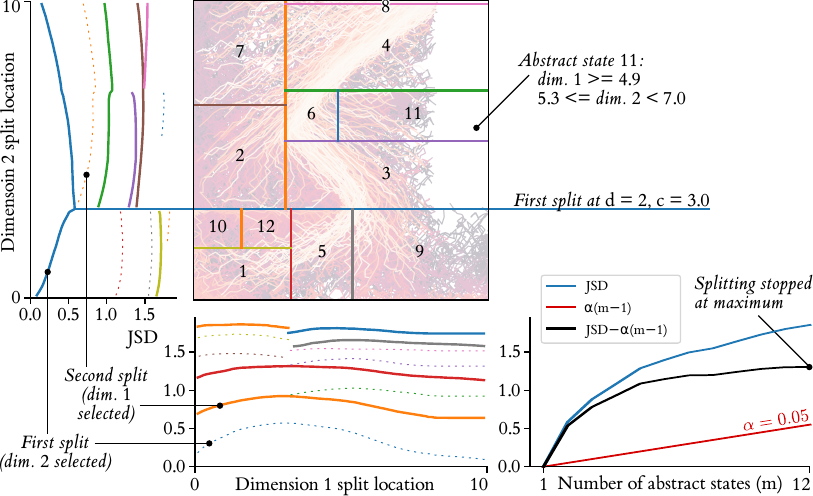}
\vspace{-0.1cm}
\caption{\small Application of $\texttt{CSA}$ to a system with $D=2$ state dimensions. $\mathcal{D}$ contains up to $200$ samples from each of $k=750$ Markov chains
and the prior $\rho$ is proportional to the sample counts. Initiating with $\mathcal{X}=\{\mathcal{S}\}$, the algorithm evaluates $\delta(\mathcal{X}\rightarrow[xdc])$ for a set of candidate splits $\mathcal{C}$ separated by intervals of $0.1$ along both dimensions. It is maximised at $d=2$, $c=3.0$ and a split is made, thereby updating $\mathcal{X}$ and increasing $m$ by $1$. Further splits are guaranteed to increase $\text{JSD}$ (note stacking of curves), but gains tend to diminish. With $\alpha=0.05$, the objective is maximised when $m=12$, and splitting terminates. The resultant set of rectangular abstract states are defined by their boundaries along both dimensions. \vspace{-0.3cm}
}
\label{fig:csa_example}
\end{figure}

The \texttt{CSA} algorithm initiates with one abstract state covering the entire space,
$\mathcal{X}=\{\mathcal{S}\}$. It then proceeds to recursively partition the space into ever-smaller hyperrectangles, at each step selecting $x$, $d$ and $c$ to greedily maximise equation \ref{eq:greedy}:
\begin{equation} \label{eq:split_criterion}
    \underset{x\in\mathcal{X}}{\text{max}}\ \underset{1\leq d\leq D}{\text{max}}\ \underset{c\in\mathcal{C}_{d}}{\text{max}}\ \delta(\mathcal{X}\rightarrow[xdc]).
\end{equation}
$\mathcal{C}_{d}$ is a finite set of candidate split thresholds along dimension $d$, defined either manually or using the data (e.g. all values occurring in $\mathcal{D}$). The algorithm terminates when all possible $\delta$ values are $\leq0$, which occurs earlier for larger values of $\alpha$. Figure \ref{fig:csa_example} provides an example of the algorithm in operation. 

\subsection{\texttt{CSTA} (\textit{Contrastive Spatiotemporal Abstraction})}

The first step in this extended algorithm is to apply an initial temporal abstraction $\mathcal{T}_\text{init}$, which should be chosen to reduce the sparsity of $J^{\mathcal{X},\mathcal{T}_\text{init}}$ to an acceptable level, but to otherwise have as many windows as possible to avoid excessive smoothing of the nonstationary dynamics (if sparsity is not an issue, the null abstraction $\mathcal{T}_\text{null}=\{[1\leq i<2],...,[k\leq i<k+1]\}$ can be used). After applying the initial temporal abstraction, $\texttt{CSA}$ is called as a subroutine, yielding a state abstraction $\mathcal{X}$ that maximises the $\text{JSD}$ between the initial windows. 

We then run a secondary stage of temporal partitioning to find another set of windows $\mathcal{T}$ (typically $|\mathcal{T}|\ll|\mathcal{T}_\text{init}|$) that preserve $\text{JSD}$ as much as possible. This is also a recursive procedure. Given an existing $\mathcal{T}$ consisting of $n$ windows, splitting the $w\in\{1,...,n\}$th window at a location $i\in\{2,...,k\}$ creates two child windows with respective boundaries $l_w,i$ and $i,u_w$. This yields an enlarged transition array with shape $(n+1)\times m\times m$, denoted by $J^{\mathcal{X},\mathcal{T}\rightarrow[wi]}$. 

\begin{figure}[h!]
\centering
\includegraphics{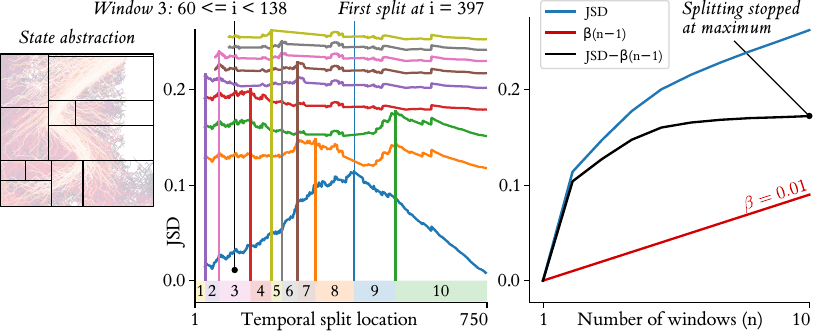}
\vspace{-0.15cm}
\caption{\small Application of the \texttt{CSTA} temporal partitioning stage to the dataset in Figure \ref{fig:csa_example}, given the $m=12$ state abstraction shown (note that $\mathcal{T}_\text{init}=\mathcal{T}_\text{null}$ was used here). Initiating with one time window, $\mathcal{T}=\{[1\leq i<k+1]\}$, the algorithm evaluates $\delta(\mathcal{T}\rightarrow[wi])$ for candidate splits $\mathcal{C}_\text{temporal}$, here being the exhaustive set $\{2,...,750\}$, excluding those that violate the minimum window width $\varepsilon=25$. It is maximised at $i=397$ and a split is made, thereby updating $\mathcal{T}$ and increasing $n$ by $1$. Further splits are guaranteed to increase $\text{JSD}$ (note stacking of curves), but gains tend to diminish. With $\beta=0.01$, the objective is maximised when $n=10$, and splitting terminates.
} \vspace{-0.25cm}
\label{fig:csta_example}
\end{figure}

Initiating with one window covering the entire chain sequence, $\mathcal{T}=\{[1\leq i<k+1]\}$, \texttt{CSTA} iteratively adds windows by splitting, at each step selecting $w$ and $i$ to greedily maximise performance according to equation \ref{eq:objective_with_windows}:
$$\underset{1\leq w\leq n}{\text{max}}\ \underset{i\in\mathcal{C}_\text{temporal}}{\text{max}}\ \delta(\mathcal{T}\rightarrow[wi]),\ \text{where}\ \ \delta(\mathcal{T}\rightarrow[wi])=$$
\vspace{-0.45cm}
\begin{equation} \label{eq:split_criterion_temporal}
\text{JSD}(J^{\mathcal{X},\mathcal{T}\rightarrow[wi]}|\rho^{\mathcal{T}\rightarrow[wi]})-\text{JSD}(J^{\mathcal{X},\mathcal{T}}|\rho^\mathcal{T})-\beta,
\end{equation}
and $\mathcal{C}_\text{temporal}$ is a finite set of candidate temporal split thresholds. In our implementation, we exclude from $\mathcal{C}_\text{temporal}$ any threshold that would violate a minimum window width $\varepsilon$. The algorithm terminates as soon as either this condition leaves no valid thresholds, or all possible $\delta$ values are $\leq 0$, the latter of which occurs earlier for larger values of $\beta$. Algorithm \ref{alg:csta} contains pseudocode for \texttt{CSTA}, and Figure \ref{fig:csta_example} provides an example of the temporal partitioning stage.

\begin{algorithm}[h]
\small
\caption{\small \texttt{CSTA} (annotated copy in Appendix B)}
\label{alg:csta}
\begin{spacing}{0.975}
\begin{algorithmic}
\STATE \textbf{Inputs}: Data $\mathcal{D}$, prior $\rho$, hyperparams $\mathcal{T}_\text{init},\alpha,\beta,\varepsilon,\mathcal{C},\mathcal{C}_\text{temporal}$ 
\STATE \textbf{Initialise}: $\mathcal{X}=space(\mathcal{D})$;
\STATE $J^{\mathcal{X},\mathcal{T}_\text{init}}=joint\_probs(\mathcal{D},\mathcal{X},\mathcal{T}_\text{init},\rho)$ (eq. 1, eq. 5)
\STATE \textbf{while} true \textbf{do} \textcolor{orange}{\textsc{(state abstraction)}}
\STATE \quad \textbf{for} $x\in\mathcal{X},1\leq d\leq dim(\mathcal{D}),c\in valid(\mathcal{C}_d,x)$ \textbf{do}
\STATE \quad \quad $J^{\mathcal{X}\rightarrow[xdc],\mathcal{T}_\text{init}}=split\_state\_probs(J^{\mathcal{X},\mathcal{T}_\text{init}},x,d,c)$
\STATE \quad \quad Compute $\delta(\mathcal{X}\rightarrow[xdc])$ via eq. 7
\STATE \quad $[xdc]^*=\text{argmax}\ \delta(\mathcal{X}\rightarrow\cdot)$; \textbf{if} $\delta(\mathcal{X}\rightarrow[xdc]^*)\leq 0$: \textbf{break}
\STATE \quad $\mathcal{X}=split\_state(\mathcal{X},[xdc]^*); J^{\mathcal{X},\mathcal{T}_\text{init}}=J^{\mathcal{X}\rightarrow[xdc]^*,\mathcal{T}_\text{init}}$
\STATE \textbf{Initialise}: $\mathcal{T}=all\_i(\mathcal{D});J^{\mathcal{X},\mathcal{T}}=joint\_probs(\mathcal{D},\mathcal{X},\mathcal{T},\rho)$
\STATE \textbf{while} true \textbf{do} \textcolor{orange}{\textsc{(temporal abstraction)}}
\STATE \quad \textbf{for} $1\leq w\leq|\mathcal{T}|,i\in valid(\mathcal{C}_\text{temporal},w,\varepsilon)$ \textbf{do}
\STATE \quad \quad $J^{\mathcal{X},\mathcal{T}\rightarrow[wi]}=split\_window\_probs(J^{\mathcal{X},\mathcal{T}},w,i,\rho)$
\STATE \quad \quad Compute $\delta(\mathcal{T}\rightarrow[wi])$ via eq. 9
\STATE \quad $[wi]^*=\text{argmax}\ \delta(\mathcal{T}\rightarrow\cdot)$; \textbf{if} $\delta(\mathcal{T}\rightarrow[wi]^*)\leq 0$: \textbf{break}
\STATE \quad $\mathcal{T}=split\_window(\mathcal{T},[wi]^*); J^{\mathcal{X},\mathcal{T}}=J^{\mathcal{X},\mathcal{T}\rightarrow[wi]^*}$
\end{algorithmic}
\end{spacing}
\end{algorithm}

\newpage
\section{LEARNING AGENT EXPERIMENTS}

Through Sections \ref{sec:problem}-\ref{sec:algos}, we have remained context-agnostic by referring to $\mathcal{P}$ as a generic sequence of Markov chains over a common state space. We now return to the motivating use case given in Section \ref{sec:intro}:
learning agents for control. 

Many control problems can be formalised as Markov decision processes (MDPs), which extend Markov chains with transition probabilities that depend on agent actions $a\in\mathcal{A}$. Given an action-selection policy $\pi_i:\mathcal{S}\rightarrow\Delta(\mathcal{A})$, we write this as $P_i(s'|s)=\sum_{a\in\mathcal{A}}\pi_i(a|s)T(s'|s,a),\ \forall s\in\mathcal{S},\ \forall s'\in\mathcal{S}$,
where $T:\mathcal{S}\times\mathcal{A}\rightarrow\Delta(\mathcal{S})$ is a fixed dynamics function.
Regardless of the agent's learning algorithm or objective (our method is agnostic to both), online 
learning yields a temporal sequence of policies, $\Pi=(\pi_1,...,\pi_k)$, 
and in turn, a sequence of Markov chains $\mathcal{P}=(P_1,...,P_k)$. The history of experience gathered by the agent during learning can be represented by a transition dataset $\mathcal{D}$ comprising elements of the form $(i,s,s')$, where $i$ is the index of the current policy and $s,s'$ is a pair of successive states sampled from its induced 
Markov chain.\footnote{We handle episodic MDPs by adding an absorbing ``pseudo-state" $te$ which is transitioned to when an episode terminates. 
} 
This is precisely the empirical setup assumed in Section \ref{sec:problem}, so we can apply contrastive spatiotemporal abstraction to summarise and compare the state transition dynamics of the agent during learning. Since the number of policy update steps $k$ can be very large for modern learning algorithms, and the incremental nature of learning tends to produce significant similarities over short timescales, temporal abstraction is both suitable and necessary.

In this section, we apply our method to construct interpretable contrastive summaries of the learning of deep RL agents in two continuous episodic MDPs: a 2D maze and a lunar landing problem \cite{gym}. In both cases, the agents take one policy update step per episode. Further details of the MDPs and RL agents are given in Appendix C.

\subsection{Maze Environment (Figure \ref{fig:results_holonav})}

In this MDP, shown in (a), the agent controls the position 
$s=[x,y]\in[0,10]^2$ 
of a marker via its velocity
$a\in[-.25,.25]^2$.
It attains $+100$ reward for reaching a goal and $-100$ for entering a penalty region, and walls at $y=3$ and $y=7$ block transverse motion. Episodes terminate on entering the goal or penalty, or after $200$ timesteps. The RL algorithm has the objective of maximising reward, so converges over $k=750$ episodes to a policy that moves to the goal while avoiding the walls, as shown in the transition data plot (b). Figure \ref{fig:csa_example} uses this dataset, and shows that our algorithm makes its first and third splits at $y=3$ and $y=7$ respectively, which align exactly with the walls. The sharp peaks in $\text{JSD}$ around these locations, visible in the left sub-plot, suggest that this is no coincidence, and that the changing ability of the agent to pass each wall is a major source of divergence in the policy sequence. Other splits have similarly intuitive origins, with abstract states $6$ and $8$ respectively covering a critical junction at the upper wall where the agent has to learn a sharp right turn, and part of the upper maze boundary along which the agent learns to ``slide" to the goal near the end of learning. Returning to Figure \ref{fig:results_holonav}, (e) provides an alternative visualisation of the state abstraction as a tree diagram, showing how the $m=12$ abstract states are hierarchically defined by the splits.

The temporal abstraction stage, whose details are the subject of Figure \ref{fig:csta_example}, yields $n=10$ windows, which are overlaid on the agent's learning curve (mean reward over a $10$-episode radius) in Figure \ref{fig:results_holonav} (d). Notice the alignment between the window boundaries and various jumps and direction changes in the learning curve; this provides encouragement that the $\text{JSD}$-based split criterion is effective for partitioning learning into periods of qualitatively similar behaviour. 


\begin{figure*}[t]
\centering
\includegraphics[width=\textwidth]{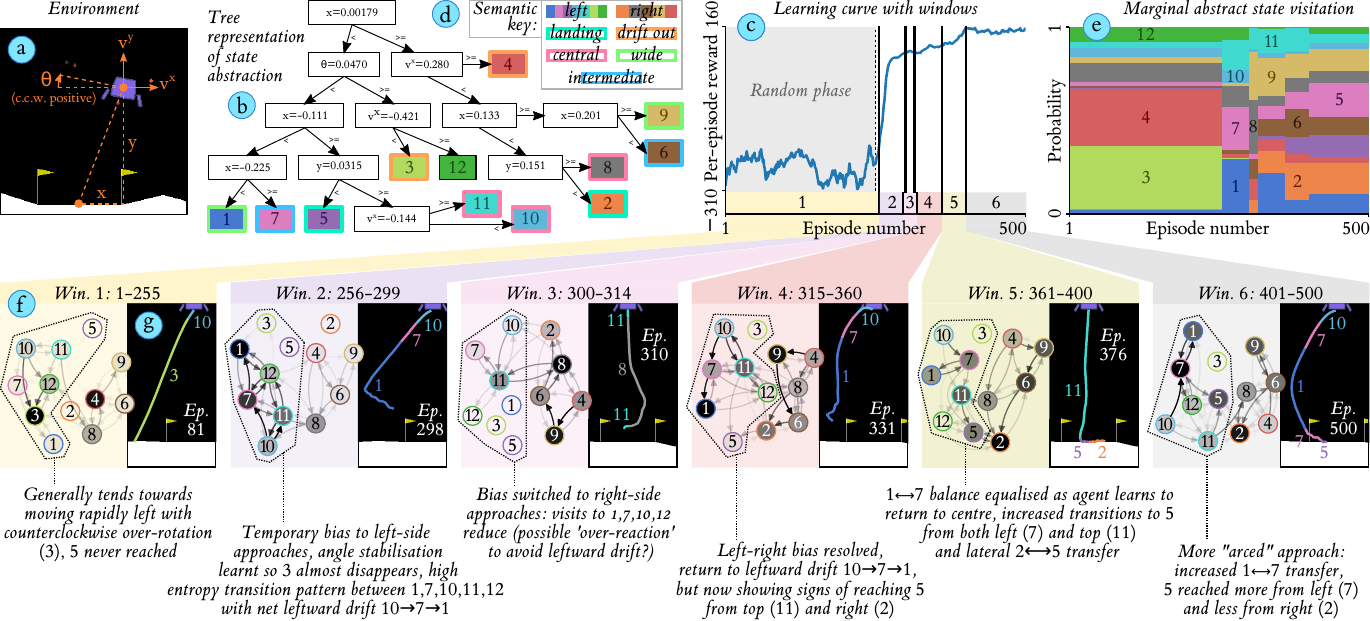}
\vspace*{-0.54cm}
\caption{{\small 
Abstraction results and analysis for an RL agent in LunarLander. Hyperparameter values: prior $\rho$ proportional to sample counts; $\mathcal{T}_\text{init}=\mathcal{T}_\text{null}$; $\alpha=0.05$; $\beta=0.01$; $\varepsilon=15$; $\mathcal{C}_d$ set to percentiles of data along each dimension $d$; $\mathcal{C}_\text{temporal}$ set to exhaustive set $\{2,...,500\}$.
\vspace*{-0.35cm}
}}
\label{fig:results_lunarlander}
\end{figure*}

For windows $3$ and $10$, we examine the abstract state dynamics in detail using transition graphs in Figure \ref{fig:results_holonav} (f), where node/edge opacities scale with state visitation and joint transition probabilities respectively. From these, we can derive contrastive interpretations of the agent's learning, such as:
\begin{itemize}
\item The sparser and less symmetric graph for window $10$ indicates a general trend towards goal-directed behaviour.
\item Transitions providing progress to the goal (e.g. $3\rightarrow6$) become more common, while those leading to corners or dead-ends (e.g. $9\rightarrow5$) occur less frequently.
\item Abstract state $1$ sees both a decrease in visitation and a focusing of outbound transitions to state $5$ only. Inbound transitions to state $1$ disappear entirely; in window $10$, the agent only visits the state if initialised there.
\end{itemize}
Stacked bar charts in Figure \ref{fig:results_holonav} (g, h) give a complementary view on the changing dynamics across all windows. In (g), bar heights reflect the marginal visitation distribution, and in (h) they represent the conditional outbound probabilities from abstract state $6$ only. They provide further contrastive insight:
\begin{itemize}
\item Initially approximately uniform, marginal visitation gradually skews towards abstract state $4$ and away from $2,7$ and $9$. Visits to state $3$ remain largely unchanged.
\item Aberrations include two ``waves" of visitation to state $1$, in which the agent fails to exit the bottom-left corner.
\item The outbound $6\rightarrow4$ transition comes to dominate, indicating a learnt ability to turn past the upper wall.
\item The $6\rightarrow11$ transition briefly spikes in window $6$; at this time the agent often becomes stuck below the wall.
\end{itemize}

\subsection{LunarLander Environment (Figure \ref{fig:results_lunarlander})}

In this MDP, Figure \ref{fig:results_lunarlander} (a), the agent uses two thrusters to land a craft between flags, with $\pm100$ reward for landing/crashing, and shaping reward promoting smooth flight. Episodes end on landing/crashing, or after $400$ timesteps. To sense-check the temporal abstraction, we run a random policy for $250$ episodes then enable the RL algorithm for a further $250$.

Contrastive spatiotemporal abstraction yields $12$ abstract states (b) and $6$ time windows (c). The inherent symmetry of the MDP is reflected in splits along the horizontal state dimension $x$, with the first close to the centreline 
and others near $\pm 0.1$ and $\pm0.2$ (note: flags are at $\pm 0.2$). $8$ out of $11$ splits concern the agent's behaviour in $x$ (including velocity $v^x$), suggesting most temporal contrasts occur along this dimension: vertical dynamics are largely gravity-driven, thus remain constant over learning. Exceptions are the splits along $y$ to create abstract states $2$ and $5$, corresponding to being low and central (i.e. correct landing) on the right/left respectively. These and other linguistic labels are given in a \textit{semantic key} (d), which aids us in interpreting abstract state dynamics.

Turning to the temporal abstraction (c), we see that all splits lie in the RL phase, before the agent converges to a stable reward of $\approx 150$. This passes the aforementioned sense-check, and aligns with theory: $\text{JSD}$ is maximised by splitting at times when transition probabilities are changing; splitting a stationary window gives zero $\text{JSD}$ gain in expectation.

(e) shows marginal abstract state visitation. In window $1$ most time is spent drifting out from the centre (states $3,4$) but this is rapidly unlearnt. Visits to the landing states $2,5$ increase; others oscillate in prevalence before convergence.

A richer narrative emerges from the transition graphs for all six windows (f). We focus on states on the left of the state space, enclosed in dotted lines. The trends are complex and nonlinear, as expected from a stochastic learning agent, but our annotations tell the story of over-rotated leftward motion in state $3$ giving way to a slower leftward drift $10\rightarrow 7\rightarrow 1$, punctuated by a brief abandonment of left-side approaches, and finally a convergence to landing state $5$ with a preference for arced approaches via states $1$ and $7$. Such insight would typically require a laborious review of the learning history, but is automatically mined by contrastive abstraction.

The preceding interpretation is reinforced by visualising the \textit{prototype} episode (g) for each window $w$, that which is most representative of the window's dynamics according to the mean log likelihood over the conditional transition distribution: \smash{$\text{argmax}_{l_w\leq i<u_w}[\frac{1}{\text{len}(i)}\sum_{t=0}^{\text{len}(i)-1}\log P^{\mathcal{X},\mathcal{T}}_{w,x_{i,t},x_{i,t+1}}]$}. In Appendix D we show how a similar calculation allows us to dynamically assess the representativeness of an ongoing episode with respect to each window, and in turn provides a mechanism for \textit{counterfactual review} of critical transitions.

\vspace{-0.1cm}
\section{CONCLUSION}
Contrastive spatiotemporal abstraction
aggregates transition data to generate an interpretable summary of the salient points of contrast within a Markov chain sequence. The technique is agnostic to the data's generative origins, making it widely applicable, although we have 
evaluated it in the context of learning agents in continuous MDPs. In this context, worthwhile extensions would be to ``seed" the state abstraction with regions of interest such as goals and penalties, augment the state space with action and reward information, or adapt our algorithms to perform online abstraction from streaming data during agent learning. It is natural to consider generalisations to non-hyperrectangular abstract states, and to fuzzy time windows that more faithfully capture smooth temporal trends, although this may hamper interpretability. There are also many possibilities for the graphical and textual representation of abstract dynamics models, beyond those explored here. However it is deployed or extended, we hope our method will be a valuable contribution to the agent interpretability ecosystem. 

\newpage
\balance
\bibliographystyle{apalike} 
\bibliography{bibliography}

\clearpage

\onecolumn

\section*{APPENDIX A. SUBLINEARITY OF THE JENSEN-SHANNON DIVERGENCE WITH $m$ AND $n$}

\subsection*{Theoretical Analysis for $m$}

For reference, the form of the Jensen-Shannon divergence given in the main paper is
$$\text{JSD}(J^\mathcal{X}|\rho)=-\sum_{i=1}^k\rho_i\mathcal{H}(J_i^\mathcal{X})+\mathcal{H}\Big(\sum_{i=1}^k\rho_iJ_i^\mathcal{X}\Big),$$
which can be expanded as follows:
$$=\left(\sum_{i=1}^k\rho_i\sum_{x\in\mathcal{X}}\sum_{x'\in\mathcal{X}}J_{i,x,x'}^\mathcal{X}\log J_{i,x,x'}^\mathcal{X}\right)-\left(\sum_{x\in\mathcal{X}}\sum_{x'\in\mathcal{X}}\Big(\sum_{i=1}^k\rho_iJ_{i,x,x'}^\mathcal{X}\Big)\log\Big( \sum_{j=1}^k\rho_jJ_{j,x,x'}^\mathcal{X}\Big)\right).
$$

Consider the scaling of $\text{JSD}$ with $m$ under a \textit{binary partition} operation, which involves replacing a randomly-selected abstract state $x_0\in\mathcal{X}$ with two others $x_1$ and $x_2$, subject to the following conservation equations for all $i\in\{1,...,k\}$:
$$J_{i,x_0,x_p}=J_{i,x_1,x_p}+J_{i,x_2,x_p},\ \ \forall x_p\in\mathcal{X}^-\ \ \text{(Outbound transitions conserved)};$$
$$J_{i,x_p,x_0}=J_{i,x_p,x_1}+J_{i,x_p,x_2},\ \ \forall x_p\in\mathcal{X}^-\ \ \text{(Inbound transitions conserved)};$$
$$J_{i,x_0,x_0}=J_{i,x_1,x_1}+J_{i,x_1,x_2}+J_{i,x_2,x_1}+J_{i,x_2,x_2}\ \ \text{(Internal transitions conserved)},$$
where $\mathcal{X}^-$ denotes all abstract states except $x_0$, $x_1$ and $x_2$ themselves, and we have dropped the superscripted $\mathcal{X}$ from $J^\mathcal{X}_{\cdot,\cdot,\cdot}$ to reduce visual clutter. All other transition probabilities (i.e. between the members of $\mathcal{X}^-$) remain unchanged. This scaling analysis is of interest because the algorithm presented in Section 3 of the main paper employs a binary partitioning approach.

Let $\delta_\text{JSD}$ denote the difference between the $\text{JSD}$ values before and after the partition is made. Most terms of this difference cancel, and we are left with
$$\delta_\text{JSD}=\sum_{x_p\in\mathcal{X}^-}\Big(z_{p,1}+z_{1,p}+z_{p,2}+z_{2,p}-z_{p,0}-z_{0,p}\Big)+z_{1,1}+z_{1,2}+z_{2,1}+z_{2,2}-z_{0,0},$$
where 
$$z_{p,q}=\Big(\sum_{i=1}^k\rho_iJ_{i,x_p,x_q}\log J_{i,x_p,x_q}\Big)\\-\Big(\sum_{i=1}^k\rho_iJ_{i,x_p,x_q}\Big)\log\Big( \sum_{i=1}^k\rho_iJ_{i,x_p,x_q}\Big).$$

A general analysis of $\delta_\text{JSD}$ is not straightforward, so here we only consider the special case of $k=2$ Markov chains with prior weighting $\rho_1=\rho_2=\frac{1}{2}$. As a lower bound, if the underlying state transition probabilities in $\mathcal{S}$ are identical for chains $1$ and $2$, it is easy to show that all terms cancel and $\delta_\text{JSD}=0$. More interestingly, we can upper bound the $\text{JSD}$ difference by considering the ``perfect partitioning" case when chains $1$ and $2$ have the \textit{same} inbound and outbound transition probabilities for $x_0$ (i.e. $J_{1,x_p,x_0}=J_{2,x_p,x_0}$ and $J_{1,x_0,x_p}=J_{2,x_0,x_p}$, $\forall x_p\in\mathcal{X}^-\cup\{x_0\}$), but chain $1$ \textit{only} visits $x_1$ and chain $2$ \textit{only} visits $x_2$. This leads to the following simplification of the terms $z_{p,1}+z_{p,2}-z_{p,0}$ for each $x_p\in\mathcal{X}^-$:
\begin{align*}
z_{p,1}+z_{p,2}=2\Big(\frac{1}{2}J_{\cdot,x_p,x_0}\log J_{\cdot,x_p,x_0}-\frac{1}{2}J_{\cdot,x_p,x_0}\log\frac{1}{2}J_{\cdot,x_p,x_0}\Big);\\
z_{p,0}=\frac{2}{2}J_{\cdot,x_p,x_0}\log J_{\cdot,x_p,x_0}-\frac{2}{2}J_{\cdot,x_p,x_0}\log\frac{2}{2}J_{\cdot,x_p,x_0}=0;\\
z_{p,1}+z_{p,2}-z_{p,0}=J_{\cdot,x_p,x_0}\log\frac{\cancel{J_{\cdot,x_p,x_0}}}{\frac{1}{2}\cancel{J_{\cdot,x_p,x_0}}}-0=J_{\cdot,x_p,x_0}\log 2,
\end{align*}
where the subscripted placeholder ``$\cdot$" indicates that this joint probability is the same for both chains. An equivalent simplification can be derived for $z_{1,p}+z_{2,p}-z_{0,p}$, and also for $z_{1,1}+z_{2,2}-z_{0,0}$. We also know that $z_{1,2}=z_{2,1}=0$, because (by assumption) neither of the two chains ever visits both $x_1$ and $x_2$. Substituting back into the equation for $\delta_\text{JSD}$, we have
$$
\delta_\text{JSD}=\sum_{x_p\in\mathcal{X}^-}\Big(J_{\cdot,x_p,x_0}\log 2+J_{\cdot,x_0,x_p}\log 2\Big)+J_{\cdot,x_0,x_0}\log 2
=M_{\cdot,x_0}\log2+\sum_{x_p\in\mathcal{X}^-}J_{\cdot,x_p,x_0}\log 2,
$$
where $M_{\cdot,x_0}=\sum_{x_p\in\mathcal{X}}J_{\cdot,x_0,x_p}$ is the marginal visitation probability for $x_0$, obtained by summing over all joint outbound transition probabilities. This quantity is the same for both chains.

What can we say about the behaviour of $\delta_\text{JSD}$ in expectation? For any Markov chain $i$, $\sum_{x\in\mathcal{X}}M_{i,x}=1$, so if $x_0$ was selected for partitioning with uniform random probability, $\mathbb{E}_{x_0}[M_{\cdot,x_0}]=\frac{1}{m}$ (where $m$ is the number of abstract states \textit{before} the partition is made). We can also bring the expectation of the second term inside the sum:
$$
\mathbb{E}_{x_0}[\delta_\text{JSD}]=\mathbb{E}_{x_0}[M_{\cdot,x_0}]\log 2+\sum_{x_p\in\mathcal{X}^-}\mathbb{E}_{x_0}[J_{\cdot,x_p,x_0}]\log 2
=\Big(\frac{1}{m}+\sum_{x_p\in\mathcal{X}^-}\mathbb{E}_{x_0}[J_{\cdot,x_p,x_0}]\Big)\log 2.
$$

We cannot say much more about the second term in general, aside from that it will depend on the amount of ``inertia" in the Markov chains: the proportion of timesteps in which they remain in the same abstract state as opposed to transitioning elsewhere. If inertia is very high, the first term of our expression for $\mathbb{E}_{x_0}[\delta_\text{JSD}]$ will dominate, in which case
$$\mathbb{E}_{x_0}[\delta_\text{JSD}]=\frac{\log 2}{m}\implies\mathbb{E}[\text{JSD}]=\log 2\log m,$$
since $\frac{d}{dy}(a\log y)=\frac{a}{y}$. This result suggests that in the high-inertia context, $\text{JSD}$ is approximately logarithmic in $m$. Finally, as a looser upper bound, we can say for certain that $J_{\cdot,x_p,x_0}\leq M_{\cdot,x_p}$, so $\mathbb{E}_{x_0}[J_{\cdot,x_p,x_0}]\leq\frac{1}{m},\ \forall x_p\in\mathcal{X}^-$. Therefore,
$$\mathbb{E}_{x_0}[\delta_\text{JSD}]\leq\Big(\frac{1}{m}+\sum_{x_p\in\mathcal{X}^-}\frac{1}{m}\Big)\log2=\Big(\cancel{\frac{1}{m}}+\frac{m-\cancel{1}}{m}\Big)\log2=\log2\implies\mathbb{E}[\text{JSD}]\leq m\log 2.$$
since $\frac{d}{dy}(ay)=a$. In this (extremely generous) upper bound case, $\text{JSD}$ is linear in $m$. As this case will be realised very rarely in practice, it is reasonable to say that the scaling will almost always be sublinear, and more tentatively, that it is approximately logarithmic.

\subsection*{Theoretical Analysis for $n$}

A more general analysis is possible for $n$. When temporal abstraction is applied to aggregate $k$ temporally-ordered Markov chains into $n$ windows, the Jensen-Shannon divergence is given by
$$\text{JSD}(J^{\mathcal{X},\mathcal{T}}|\rho^\mathcal{T})=\left(\sum_{w=1}^n\rho_w^\mathcal{T}\sum_{x\in\mathcal{X}}\sum_{x'\in\mathcal{X}}J_{w,x,x'}^{\mathcal{X},\mathcal{T}}\log J_{w,x,x'}^{\mathcal{X},\mathcal{T}}\right)-\left(\sum_{x\in\mathcal{X}}\sum_{x'\in\mathcal{X}}\Big(\sum_{w=1}^n\rho_w^\mathcal{T}J_{w,x,x'}^{\mathcal{X},\mathcal{T}}\Big)\log\Big( \sum_{v=1}^n\rho_v^\mathcal{T}J_{w,x,x'}^{\mathcal{X},\mathcal{T}}\Big)\right).
$$

Consider the replacement of a randomly-selected time window $w_0\in\{1,...,n\}$ with two others $w_1$ and $w_2$, subject to the following conservation equations for all $x\in\mathcal{X},x'\in\mathcal{X}$:
$$\rho_{w_0}=\rho_{w_1}+\rho_{w_2} \ \ \text{(Sum of prior weights conserved)};$$
$$\rho_{w_0}J_{w_0,x,x'}=\rho_{w_1}J_{w_1,x,x'}+\rho_{w_2}J_{w_2,x,x'} \ \ \text{(Joint probabilities are prior-weighted average)},$$
where we have dropped the superscripted $\mathcal{X}$ and $\mathcal{T}$ from both $\rho^\mathcal{T}$ and $J^{\mathcal{X},\mathcal{T}}_{\cdot,\cdot,\cdot}$ to reduce visual clutter. Note that this form of conservation results from partitioning a window in two. 

Again taking $\delta_\text{JSD}$ to be the difference between the $\text{JSD}$ values before and after the partitioning, all of the second term (of the form given above), and most components of the first term, cancel out. We are left with
$$
\delta_\text{JSD}=\sum_{x\in\mathcal{X}}\sum_{x'\in\mathcal{X}}\Big(\rho_{w_1}J_{w_1,x,x'}\log J_{w_1,x,x'}+\rho_{w_2}J_{w_2,x,x'}\log J_{w_2,x,x'}-\rho_{w_0}J_{w_0,x,x'}\log J_{w_0,x,x'}\Big).
$$

Applying the conservation equations and multiplying and dividing by $\rho_{w_0}$, we can rewrite this as
\begin{align*}
\delta_\text{JSD}=\rho_{w_0}\sum_{x\in\mathcal{X}}\sum_{x'\in\mathcal{X}}\Big(\frac{\rho_{w_1}}{\rho_{w_0}}J_{w_1,x,x'}\log J_{w_1,x,x'}+\frac{\rho_{w_2}}{\rho_{w_0}}J_{w_2,x,x'}\log J_{w_2,x,x'}\\
-\Big(\frac{\rho_{w_1}}{\rho_{w_0}}J_{w_1,x,x'}+\frac{\rho_{w_2}}{\rho_{w_0}}J_{w_2,x,x'}\Big)\log \Big(\frac{\rho_{w_1}}{\rho_{w_0}}J_{w_1,x,x'}+\frac{\rho_{w_2}}{\rho_{w_0}}J_{w_2,x,x'}\Big)\Big).
\end{align*}

By inspection, everything aside from the leading $\rho_{w_0}$ exactly matches the definition of the $\text{JSD}$ itself, calculated pairwise between windows $w_1$ and $w_2$, and using renormalised weights:
$$\delta_\text{JSD}=\rho_{w_0}\ \text{JSD}(\ [J_{w_1},J_{w_2}]\ |\ [\frac{\rho_{w_1}}{\rho_{w_0}},\frac{\rho_{w_2}}{\rho_{w_0}}]\ ).$$

$\delta_\text{JSD}$ is lower-bounded by zero in the case when $w_1$ and $w_2$ have identical transition probabilities in $\mathcal{X}$ (since pairwise $\text{JSD}$ will be zero). As an upper bound, we again consider a ``perfect partitioning" case where the two new windows have \textit{no} abstract state transitions in common (i.e. $J_{w_1,x,x'}>0\implies J_{w_2,x,x'}=0$ and $J_{w_2,x,x'}>0\implies J_{w_1,x,x'}=0$, $\forall x\in\mathcal{X},x'\in\mathcal{X}$), and equal prior weighting $\rho_{w_1}=\rho_{w_2}=\frac{\rho_{w_0}}{2}$. Returning to the expanded form of $\delta_\text{JSD}$ and substituting in these relations, we obtain
\begin{align*}
\delta_\text{JSD}=\rho_{w_0}\sum_{x\in\mathcal{X}}\sum_{x'\in\mathcal{X}}\Big(\frac{1}{2}J_{\cdot,x,x'}\log J_{\cdot,x,x'}-\Big(\frac{1}{2}J_{\cdot,x,x'}\Big)\log \Big(\frac{1}{2}J_{\cdot,x,x'}\Big)\Big)
\\=\frac{\rho_{w_0}}{2}\sum_{x\in\mathcal{X}}\sum_{x'\in\mathcal{X}}J_{\cdot,x,x'}\log \frac{\cancel{J_{\cdot,x,x'}}}{\frac{1}{2}\cancel{J_{\cdot,x,x'}}}=\frac{\rho_{w_0}\log 2}{2}\cancelto{2}{\sum_{x\in\mathcal{X}}\sum_{x'\in\mathcal{X}}J_{\cdot,x,x'}}=\rho_{w_0}\log 2,
\end{align*}
where the subscripted placeholder ``$\cdot$" now stands for \textit{either} $w_1$ or $w_2$; whichever of the two has nonzero transition probability for each $x,x'$ pair. In the final line above, the double sum cancels to $2$ because for each of the two windows, the sum of joint probabilities is $1$.

Finally, we know that $\sum_{w=1}^n\rho_w=1$, so if we assume that $w_0$ was selected uniform-randomly, $\mathbb{E}_{w_0}[\rho_{w_0}]=\frac{1}{n}$ (where $n$ is the number of chains \textit{before} the partition is made). Therefore,
$$\mathbb{E}_{x_0}[\delta_\text{JSD}]=\frac{\log 2}{n}\implies\mathbb{E}[\text{JSD}]=\log 2\log n,$$
since $\frac{d}{dy}(a\log y)=\frac{a}{y}$. We can thus conclude from this upper-bound analysis that $\text{JSD}$ is reliably sublinear in $n$.

\subsection*{Empirical Validation}

As a simple empirical validation of the preceding results, we generated datasets of $k=100$ random trajectories of length $T=100$ (one per ``chain'') in the 2D region $\mathcal{S}=[0,1]^2$ according to a parameterised Gaussian random walk model:
\begin{align*}
s_{i,0}\sim\text{uniform}(\mathcal{S}),\ \ \ \forall i\in\{1,...,k\};\\
s_{i,t}=\text{clip}(s_{i,t-1}+\eta,\mathcal{S})\ \ \ \text{where}\ \ \ \eta\sim\mathcal{N}\left(\left[\begin{array}{cc}v\sin(\frac{3\pi i}{2k})&v\cos(\frac{3\pi i}{2k})\end{array}\right],\left[\begin{array}{cc}\sigma&0\\0&\sigma\end{array}\right]\right),\ \ \ \forall i\in\{1,...,k\},\ \forall t\in\{1,...,T-1\}.
\end{align*}
The $\text{clip}(\cdot)$ function clips states into the $[0,1]^2$ region. In each trajectory, Gaussian noise with standard deviation $\sigma$ is added to a mean velocity vector with magnitude $v$, which is gradually rotated through $270^\circ$ over the sequence $i\in\{1,...,k\}$.

 We generated four datasets with various arbitrarily-chosen values of $v$ and $\sigma$. 
For each dataset, we performed recursive axis-aligned partitioning of the region up to a maximum abstraction size $m$ and measured the resultant $\text{JSD}$. To select split points we used both a random strategy and the \texttt{CSTA} algorithm presented in Section 3 of the main paper, which greedily maximises the $\text{JSD}$ gain on each step.

As predicted by the preceding theory, the scaling of $\text{JSD}$ is consistently sublinear in both $m$ and $n$. It is also always higher when the \texttt{CSTA} algorithm is used, compared with random splitting. Note that these datasets have an extremely simple dynamical structure; the advantage of principled split selection using \texttt{CSTA} is likely to be far greater in practical applications.

\begin{figure}[h!]
\centering
\includegraphics[width=\textwidth]{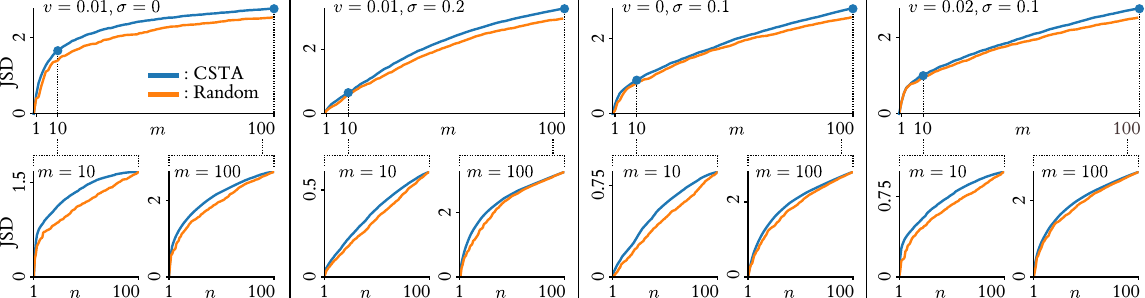}
\caption{Scaling of $\text{JSD}$ with $m$ and $n$ for random trajectories in $[0,1]^2$ with various velocity and noise parameters $v,\sigma$, and when selecting both random axis-aligned splits (orange) and those that maximise the contrastive abstraction objective (blue). Top row: scaling with $m$. Bottom row: scaling with $n$ for the specific values of $m$ shown.
}
\end{figure}

\newpage
\section*{APPENDIX B. ANNOTATED PSEUDOCODE AND SUBFUNCTION DETAILS}

For reference, the \texttt{CSTA} algorithm given in the main paper is copied below with clarifying comments in \textcolor{myblue}{blue}. The \texttt{CSA} algorithm, which excludes the temporal abstraction stage, is recovered by setting $\mathcal{T}_\text{init}=\mathcal{T}_
\text{null}$ and running lines $1$-$8$ only.

\vspace{-0.04cm}
\begin{algorithm}[h]
\small
\caption{\small \texttt{CSTA}}
\label{alg:csta}
\begin{spacing}{1.24893}
\begin{algorithmic}[1]
\STATE \textbf{Inputs}: Data $\mathcal{D}$, prior $\rho$, hyperparameters $\mathcal{T}_\text{init},\alpha,\beta,\varepsilon,\mathcal{C},\mathcal{C}_\text{temporal}$ \quad\textcolor{myblue}{\# Prior may be determined by domain-specific considerations}
\STATE \textbf{Initialise}: $\mathcal{X}=space(\mathcal{D});J^{\mathcal{X},\mathcal{T}_\text{init}}=joint\_probs(\mathcal{D},\mathcal{X},\mathcal{T}_\text{init},\rho)$ (eq. 1, eq. 5) \quad\textcolor{myblue}{\# Initial data structures required for state abstraction}
\STATE \textbf{while} true \textbf{do} \textcolor{orange}{\textsc{(state abstraction)}} \quad\textcolor{myblue}{\# Loop broken when best possible split does not increase $\alpha$-regularised objective}
\STATE \quad \textbf{for} $x\in\mathcal{X},1\leq d\leq dim(\mathcal{D}),c\in valid(\mathcal{C}_d,x)$ \textbf{do} \quad\textcolor{myblue}{\# Try splitting each extant abstract state, at each valid split point}
\STATE \quad \quad $J^{\mathcal{X}\rightarrow[xdc],\mathcal{T}_\text{init}}=split\_state\_probs(J^{\mathcal{X},\mathcal{T}_\text{init}},x,d,c)$ \quad\textcolor{myblue}{\# Create expanded joint probability array for the split defined by $x$, $d$ and $c$}
\STATE \quad \quad Compute $\delta(\mathcal{X}\rightarrow[xdc])$ via eq. 7 \quad\textcolor{myblue}{\# Evaluate the change this split causes in the $\alpha$-regularised objective}
\STATE \quad $[xdc]^*=\text{argmax}\ \delta(\mathcal{X}\rightarrow\cdot)$; \textbf{if} $\delta(\mathcal{X}\rightarrow[xdc]^*)\leq 0$: \textbf{break} \quad\textcolor{myblue}{\# Identify best split, breaking loop if objective change is non-positive}
\STATE \quad $\mathcal{X}=split\_state(\mathcal{X},[xdc]^*); J^{\mathcal{X},\mathcal{T}_\text{init}}=J^{\mathcal{X}\rightarrow[xdc]^*,\mathcal{T}_\text{init}}$ \quad\textcolor{myblue}{\# Implement selected split by updating both $\mathcal{X}$ and $J^{\mathcal{X},\mathcal{T}_\text{init}}$}
\STATE \textbf{Initialise}: $\mathcal{T}=all\_i(\mathcal{D});J^{\mathcal{X},\mathcal{T}}=joint\_probs(\mathcal{D},\mathcal{X},\mathcal{T},\rho)$ \quad\textcolor{myblue}{\# Initial data structures required for temporal abstraction}
\STATE \textbf{while} true \textbf{do} \textcolor{orange}{\textsc{(temporal abstraction)}} \quad\textcolor{myblue}{\# Loop broken when best possible split does not increase $\beta$-regularised objective}
\STATE \quad \textbf{for} $1\leq w\leq|\mathcal{T}|,i\in valid(\mathcal{C}_\text{temporal},w,\varepsilon)$ \textbf{do} \quad\textcolor{myblue}{\# Try splitting each extant time window, at each valid split point}
\STATE \quad \quad $J^{\mathcal{X},\mathcal{T}\rightarrow[wi]}=split\_window\_probs(J^{\mathcal{X},\mathcal{T}},w,i,\rho)$ \quad\textcolor{myblue}{\# Create expanded joint probability array for the split defined by $w$ and $i$}
\STATE \quad \quad Compute $\delta(\mathcal{T}\rightarrow[wi])$ via eq. 9 \quad\textcolor{myblue}{\# Evaluate the change this split causes in the $\beta$-regularised objective}
\STATE \quad $[wi]^*=\text{argmax}\ \delta(\mathcal{T}\rightarrow\cdot)$; \textbf{if} $\delta(\mathcal{T}\rightarrow[wi]^*)\leq 0$: \textbf{break} \quad\textcolor{myblue}{\# Identify best split, breaking loop if objective change is non-positive}
\STATE \quad $\mathcal{T}=split\_window(\mathcal{T},[wi]^*); J^{\mathcal{X},\mathcal{T}}=J^{\mathcal{X},\mathcal{T}\rightarrow[wi]^*}$ \quad\textcolor{myblue}{\# Implement selected split by updating both $\mathcal{T}$ and $J^{\mathcal{X},\mathcal{T}}$}
\end{algorithmic}
\end{spacing}
\end{algorithm}
\vspace{-0.26cm}

\subsection*{Subfunction Details}

\begin{itemize}
\item $space$ (line 2): Return the unit set $\{((-\infty,\infty),...,(-\infty,\infty))\}$, whose element is a $D$-tuple of identical tuples $(-\infty,\infty)$, where $D$ is the dimensionality of the state space of $\mathcal{D}$. This creates a single abstract state covering the entire state space. 
\item $joint\_probs$ (lines 2 and 9): This subfunction has two parts:
\begin{enumerate}
\item Given a dataset $\mathcal{D}$ and a state abstraction $\mathcal{X}$, use equation 1 to compute $N^\mathcal{X}$ then normalise along the second and third dimensions to obtain joint probabilities $J^\mathcal{X}$. 
\begin{itemize}
\item \textit{Note: If the data come from absorbing Markov chains, add an additional row/column to the arrays for a terminal pseudo-state $te$ and record each termination as a transition to this state.}
\end{itemize}
\item Given a $k\times m\times m$ joint probability array $J^\mathcal{X}$, an $n$-sized temporal abstraction $\mathcal{T}$, and a $k$-dimensional prior vector $\rho$, use equation 5 to compute an $n\times m\times m$ array which aggregates probabilities along the first dimension.
\end{enumerate}
\item $dim$ (line 4): Return $D$, the dimensionality of the state space of $\mathcal{D}$.
\item $valid$ (lines 4 and 11): Given a set of split points and either an abstract state or time window, return only those split points that fall within the bounds of that state/window. In the latter case, further restrict to exclude those that fall less than a value of $\varepsilon$ away from either bound.
\item $split\_state\_probs$ (line 5): Given an $n\times m\times m$ joint probability array $J^{\mathcal{X},\mathcal{T}_\text{init}}$, return an expanded $n\times(m+1)\times(m+1)$ array in which the outgoing (respectively, incoming) probabilities for abstract state $x$ are redistributed between pairs of entries on the second (third) array dimension, according to whether the start state $s$ (end state $s'$) of each underlying transition $(i,s,s')\in\mathcal{D}$ lies before or after the split point $c$ along state dimension $d$. All other probabilities are unchanged. \begin{itemize}
\item \textit{Note: In our implementation (available on request) we speed up this subfunction by maintaining a temporary expanded array for each $x,d$ pair, which is updated incrementally during a sweep over ordered split points $c$.
}
\end{itemize}
\item $split\_state$ (line 8): Given a state abstraction $\mathcal{X}$, return a copy with the tuple for abstract state $x$, denoted by $((l_{x,1},u_{x,1}),...,(l_{x,D},u_{x,D}))$, removed. It is replaced by two new tuples which are copies of the removed one except for their $d$th elements, which become $(l_{x,d},c)$ and $(c,u_{x,d})$ respectively.
\item $all\_i$ (line 9): Return the unit set $\{(l_1,u_1)\}$, where $l_1=1$, $u_1=k+1$, and $k$ is the number of Markov chains represented in $\mathcal{D}$. This creates a single time window covering the entire chain sequence.
\item $split\_window\_probs$ (line 12): Given an $n\times m\times m$ joint probability array $J^{\mathcal{X},\mathcal{T}}$, return an expanded $(n+1)\times m\times m$ array in which the probabilities for time window $w$ are redistributed between pairs of entries on the first array dimension, according to whether the chain index $j$ of each underlying transition $(j,s,s')\in\mathcal{D}$ lies before or after the temporal split point $i$. The redistribution is also weighted by the prior $\rho$. All other probabilities are unchanged.
\begin{itemize}
\item \textit{Note: As with $split\_state\_probs$, our implementation uses incrementally-updated temporary arrays to improve speed.}
\end{itemize}
\item $split\_window$ (line 15): Given a temporal abstraction $\mathcal{T}$, return a copy with the tuple for time window $w$ removed and replaced by new two tuples $(l_w,i)$, $(i,u_w)$. 
\end{itemize}

\section*{APPENDIX C. MDP AND REINFORCEMENT LEARNING DETAILS}

\subsection*{MDPs}

Our experiments were conducted in two episodic MDPs with continuous state spaces, both implemented in Python using OpenAI Gym \cite{gym}. 

\paragraph{Maze}

A simple 2D navigation task. The state dimensions are the horizontal and vertical positions of a circular black marker, $x,y\in[0,10]^2$. The objective is to move the marker to a green goal region ($x\geq 8, y\geq 7$) without entering a red penalty region ($x \geq 8, 3\leq y<7$), and while navigating around a pair of horizontal walls at $y=3$ and $y=7$. The agent's action specifies the marker's horizontal and vertical velocities $v^x,v^y$, which are bounded in $[-.25,.25]^2$ and clipped if the resultant motion vector would intersect a wall or external maze boundary. At time $t$ in an episode, the agent is given a reward of 
$$
r_t=\left\{\begin{array}{lll}+100&\text{if }x_t\geq 8\land y_t\geq 7&\text{(goal region)}\\-100&\text{if }x_t \geq 8\land 3\leq y_t<7&\text{(penalty region)}\\0&\text{otherwise.}\end{array}\right.
$$
The episode is also terminated immediately if either the goal or penalty region are entered. Otherwise, the episode terminates at a time limit $t=200$.

\paragraph{LunarLander}

A standard component of the Gym library (full name \textsc{LunarLanderContinuous-v2}), in which the objective is to guide an aerial craft to a gentle landing on a landing pad surrounded by uneven terrain. The state dimensions are the craft's horizontal and vertical positions $x,y$ and velocities $v^x,v^y$, its angle from vertical $\theta$ and angular velocity $\dot{\theta}$, and two binary contact detectors $c^l,c^r$ indicating whether the left and right landing legs are in contact with the ground. The craft is initialised in a narrow zone above the landing pad, with slightly-randomised angle and velocities. The action is a pair of throttle values for two engines: main $u^m$ and side $u^s$. For most timesteps $t$, the reward is given as
$$
r_t=\text{potential}_{t+1}-\text{potential}_{t}-0.3(u^m_t+\frac{u^s_t}{10}),
$$
where 
$$
\text{potential}_t=-100\left(\sqrt{x_t^2 + y_t^2}+\sqrt{(v^x_t)^2 + (v^y_t)^2}+|\theta_t|\right)+10\left(c^l_t+c^r_t\right).
$$
In addition, a one-off reward of $+100$ is given if the craft successfully lands on the pad, and $-100$ is given if it crashes or drifts out-of-bounds ($|x_t|\geq 1$). The distinction between a landing and a crash is based on a force analysis in an underlying rigid-body physics simulation, whose details are undocumented in the provided open source code. The episode is terminated immediately if a landing, crash or out-of-bounds event occurs. Otherwise, the episode terminates at a time limit $t=400$.

\subsection*{Reinforcement Learning Setup}

In both experiments, we used our own PyTorch implementation of the soft-actor critic (SAC) algorithm \cite{haarnoja2018soft} with a discount factor of $\gamma=0.99$, an entropy regularisation coefficient of $\alpha=0.2$, a replay buffer capacity of $20000$ samples and a minibatch size of $64$. All networks (policy and value functions) had two hidden layers of $256$ units each and were trained by backpropagation using the Adam optimiser. For each MDP we independently selected the policy network learning rate $LR_\pi$, the value network learning rate $LR_Q$, the Polyak averaging coefficient for target network updates $PC$, and the total number of training episodes $E$ as follows:

\begin{table}[h!]
\center
\renewcommand{\arraystretch}{1.5}
\begin{tabular}{c|cccc|}
\cline{2-5}
                        & \multicolumn{1}{c|}{$LR_\pi$} & \multicolumn{1}{c|}{$LR_Q$} & \multicolumn{1}{c|}{$PC$} & $E$ \\ \hline
\multicolumn{1}{|c|}{Maze} & $1e^{-4}$ & $1e^{-3}$ & $0.995$ & $750$ \\ \hline
\multicolumn{1}{|c|}{LunarLander} & $5e^{-5}$ & $5e^{-4}$ & $0.99$ & $250$* \\ \hline
\end{tabular}
\end{table}
\textit{*Note: $250$ episodes run with a random policy before enabling policy/value network updates, yielding a $500$-episode dataset.}
\\ 

In a minor departure from convention, \textbf{we modified the SAC algorithm to sample minibatches and perform network updates on the final timestep of each episode only}, thereby guaranteeing that the policy (and thus, the induced Markov chain) was stationary within each episode. This allowed us to avoid the theoretical complication of the Markov chain index $i$ changing midway through an episode, thereby making the observed transitions from chain $i$ dependent on the dynamics of the previous chain $i-1$. To compensate for the reduced number of update steps compared with per-timestep updates, each of our per-episode updates iterated over a number of minibatches equal to the length of the just-completed episode.

\newpage
\section*{APPENDIX D. POSTERIOR ANALYSIS AND COUNTERFACTUAL REVIEWS}

Given a sequence of abstract state transitions for an episode, this form of local analysis involves calculating the log posterior over the conditional transition distribution $[\log\rho_w^\mathcal{T}+\sum_{t'=0}^{t-1}\log P^{\mathcal{X},\mathcal{T}}_{w,x_t',x_{t'+1}}]$ for each window $w$ and timestep $t$. Plotting these values as a time series (with the value for the episode's true window subtracted as a baseline) indicates how representative the episode is of each window's dynamics, and how this representativeness evolves over time. In turn, this analysis can inform a \textit{counterfactual review} of critical transitions in the episode, in which we identify alternative successor abstract states that would have significantly altered the posterior probabilities for future timesteps. This provides a notion of \textit{locally-relevant contrasts} between two or more window's dynamics, which are grounded in the particular events of a given episode. 
 
\begin{figure}[h!]
\centering
\includegraphics[width=0.9\textwidth]{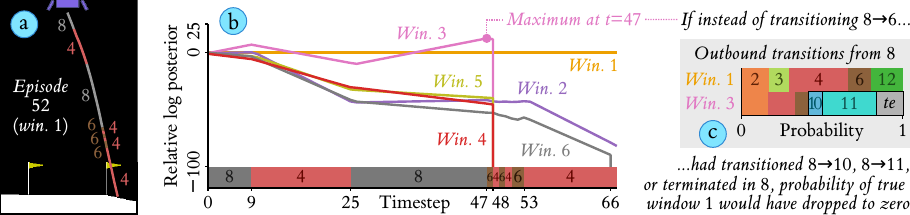}
\caption{Posterior analysis of an unsuccessful (crash) landing early in learning (a). The posterior time series (b) shows that the $6\rightarrow 4$ transition at $t=48$ eliminates windows $3$, $4$ and $5$. Windows $2$ and $6$ remain in contention until the crash terminates the episode at $t=66$, but the latter is ruled out by the final $4\rightarrow te$ transition (i.e. the agent never terminates in this abstract state near the end of learning). This leaves only windows $1$ and $2$, with the true one $1$ being more probable (recall the left-side bias in window $2$, which means this right-side episode is less representative). The relative log posterior for window $3$ is maximised at $t=47$ (just before the $8\rightarrow 6$ transition). A counterfactual review of this transition gives further insight. The outbound conditional probability plot (c) indicates that if the agent had instead transitioned to states $10$ or $11$ (both of which lie on the left half of the state space), or terminated in  state $8$, the posterior for the true window $1$ would have dropped to zero.
}
\end{figure}

\begin{figure}[h!]
\centering
\includegraphics[width=0.9\textwidth]{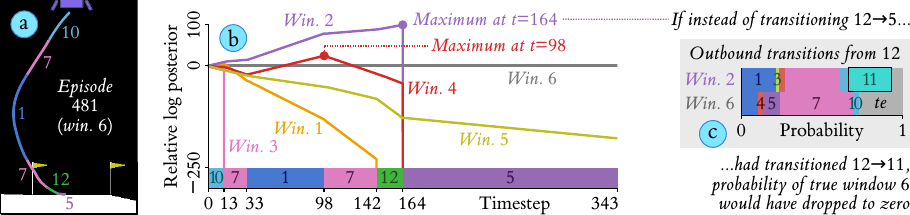}
\caption{Posterior analysis of a successful landing late in learning (a). The posterior time series (b) shows that transitioning $10\rightarrow 7$ rules out window $3$ as early as $t=13$. Up to $t=164$, the most probable window is $2$ (likely due in large part to the left-side bias in this period), but this is eliminated by the $12\rightarrow 5$ transition, leaving the true window $6$ as the winner for the remainder of the episode. A counterfactual review (c) indicates that if the agent had instead transitioned to state $11$, it would still be representative of window $2$, while eliminating window $6$. Concretely: a locally-relevant contrast between the two windows is the agent's ability to touch down (state $5$) vs stay hovering ($11$).}
\end{figure} 

\end{document}